\documentclass[draftcls,12pt,onecolumn]{IEEEtran}
\usepackage[top=.6in, bottom=.6in, left=.5125in, right=.5125in]{geometry}
\usepackage{microtype}
\usepackage{booktabs}
\usepackage{pdfpages}
\usepackage{multicol}
\usepackage{multirow,array}
\usepackage{longtable}
\usepackage{lscape}
\usepackage{epsfig}
\usepackage[T1]{fontenc}
\usepackage{graphicx}
\graphicspath{{Figures/}}
\usepackage{amssymb}
\usepackage{amsmath}
\usepackage{color}
\usepackage{epstopdf}
\usepackage{subfigure}
\usepackage{multicol}
\usepackage{multirow}
\usepackage{booktabs}
\usepackage{hyperref}       
\usepackage{url}            
\usepackage{booktabs}       
\usepackage{amsfonts}       
\usepackage{nicefrac}       
\usepackage{microtype}      
\usepackage{pdflscape}
\usepackage{graphicx}
\usepackage{subfigure}
\usepackage{amssymb} 
\usepackage{algorithm}
\usepackage{algorithmic}
\usepackage{url}
\usepackage{mathpazo}
\newcommand{\EQ}{\begin{eqnarray}}

\newcommand{\EN}{\end{eqnarray}}
\newcommand{\EQQ}{\begin{eqnarray*}}
\newcommand{\ENN}{\end{eqnarray*}}
\newcommand{\overlineray}{\begin{array} }

\newcommand{\barray}{\begin{array} }
\newcommand{\earray}{\end{array}}

\newcommand{\bremark}{\begin{remark} }
\newcommand{\eremark}{\end{remark}}
\newcommand{\btheorem}{\begin{theorem}}
\newcommand{\etheorem}{\end{theorem}}
\newcommand{\blemma}{\begin{lemma}}
\newcommand{\elemma}{\end{lemma}}
\newcommand{\bassumption}{\begin{assumption} }
\newcommand{\eassumption}{\end{assumption}}
\newcommand{\bcorollary}{\begin{corollary} }
\newcommand{\ecorollary}{\end{corollary}}
\newcommand{\bdefinition}{\begin{definition} }
\newcommand{\edefinition}{\end{definition}}
\newcommand{\bproposition}{\begin{proposition}}
\newcommand{\eproposition}{\end{proposition}}
\newcommand{\bproof}{\noindent{\it Proof: }}
\newcommand{\eproof}{\hfill\rule{2mm}{2mm}}
\newcommand{\balgorithm}{\medskip\begin{algorithm} \rm}
\newcommand{\ealgorithm}{
\end{algorithm} }
\newtheorem{remark}{\rm\bfseries Remark}
\newtheorem{corollary}{\rm\bfseries Corollary}
\newtheorem{definition}{\rm\bfseries Definition}
\newtheorem{theorem}{\rm\bfseries Theorem}
\newtheorem{lemma}{\rm\bfseries Lemma}
\newtheorem{assumption}{\rm\bfseries Assumption}
\newtheorem{proposition}{\rm\bfseries Proposition}

\definecolor{orange}{RGB}{255,127,0}

\makeatletter
\let\saved@includegraphics\includegraphics
\AtBeginDocument{\let\includegraphics\saved@includegraphics}

\makeatother
\pdfoutput=1 

\begin{document}

\title{\bf \Large DeceFL: A Principled Decentralized Federated Learning Framework}
\author{\small Ye Yuan$^{1,2,*}$, Jun Liu$^{3,*}$, Dou Jin$^{1,*}$, Zuogong Yue$^{1,*}$, Ruijuan Chen$^1$, Maolin Wang$^1$, Chuan Sun$^1$, \\ Lei Xu$^4$, Feng Hua$^2$, Xin He$^2$, Xinlei Yi$^5$, Tao Yang$^4$, Hai-Tao Zhang$^{1,2}$, Shaochun Sui$^6$, Han Ding$^2$ \thanks{$^1$School of Artificial Intelligence and Automation, Huazhong University of Science and Technology. $^2$School of Mechanical Science and Engineering, Huazhong University of Science and Technology. $^3$Department of Applied Mathematics, University of Waterloo. $^4$State Key Laboratory of Synthetical Automation for Process Industries, Northeastern University. $^5$School of Electrical Engineering and Computer Science, and Digital Futures, KTH Royal Institute of Technology. $^6$AVIC Chengdu Aircraft Industrial (Group) Co., Ltd.. $^*$Equal contributions. Email: yye@hust.edu.cn.}}

\maketitle
\begin{abstract}
Traditional machine learning relies on a centralized data pipeline, i.e., data are provided to a central server for model training. In many applications, however, data are inherently fragmented. Such a decentralized nature of these databases presents the biggest challenge for collaboration: sending all decentralized datasets to a central server raises serious privacy concerns. Although there has been a joint effort in tackling such a critical issue by proposing privacy-preserving machine learning frameworks, such as federated learning, most state-of-the-art frameworks are built still in a centralized way, in which a central client is needed for collecting and distributing model information (instead of data itself) from every other client, leading to high communication pressure and high vulnerability when there exists a failure at or attack on the central client. Here we propose a principled decentralized federated learning algorithm (DeceFL), which does not require a central client and relies only on local information transmission between clients and their neighbors, representing a fully decentralized learning framework. It has been further proven that every client reaches the global minimum with zero performance gap and achieves the same convergence rate $O(1/T)$ (where $T$ is the number of iterations in gradient descent) as centralized federated learning when the loss function is smooth and strongly convex. Finally, the proposed algorithm has been applied to a number of applications to illustrate its effectiveness for both convex and nonconvex loss functions, demonstrating its applicability to a wide range of real-world medical and industrial applications.
\end{abstract}

\clearpage
An urgent challenge for AI application today consists of the following dilemma concerning data privacy: on one hand, a large number of sophisticated algorithms have been proposed to broaden the applicability of  AI to various applications such as medicine, manufacturing and more \cite{lecun2015deep, yuan2020general, yan2020interpretable}; on the other hand, regulations such as General Data Protection Regulation (GDPR) restrict data sharing, thus limiting the performance of AI algorithms \cite{price2019privacy}. As a result, models that are trained and evaluated on a limited amount of data due to privacy could have biases \cite{degrave2021ai}. This has become a well-known bottleneck in medical AI \cite{roberts2021common}.

Promising privacy-preserving methods such as federated learning can help
maintain the performance of AI algorithms, while preserving the data
stored locally \cite{konen2015federated}. Inspired by this, there has
been a surge of interests in both the theory and applications of
federated learning \cite{Yang2019}. Federated averaging (FedAvg), the
leading algorithm in the field of federated learning, was proposed in
2016 by researchers at Google \cite{konen2016federated,
  pmlr_mcmahan2017}. Through crowded efforts and comprehensive surveys
\cite{DBLP2019Bonawitz, gu2019distributed, kairouz2019advances,
  Li2020SPM}, widely used federated learning methods were established,
the challenges and related applications of federated learning were
introduced, and a large number of valuable research directions were
outlined. A notable example has been demonstrated in a report by Kaissis
et al. \cite{kaissis2021end}, in which a convolutional neural network
was trained over the public Internet with encryption from medical images
using a secure federated learning framework.

Despite these breakthrough, classical federated learning algorithms have a major drawback:  the need for a central client, which could cause privacy, communication, computation, and resilience issues \cite{lian2017can}. Much effort, therefore, has been invested to reduce the communication and computational complexity of centralized federated learning algorithms. In order to deal with the challenge of system constraints, the authors of \cite{tang2020communicationefficient} applied sparse technology to reduce the communication and computing costs in the training process. In addition, the authors in \cite{Wang2019IEEE} proposed an optimal tradeoff control algorithm between local update and global parameter aggregation in the resource constrained system. Recently, federated schemes can be extended to a time-varying centralized scheme, where a changing leader is selected based on certain rules, which is a firm step to full decentralization \cite{2021Nature_SL}.

Here we propose the first principled decentralized federated learning algorithm, in which each client is guaranteed to achieve the same performance as the centralized algorithm in terms of training/test accuracy, when the global objective function is smooth and strongly convex. Empirical experiments have also demonstrated that the same claim holds for nonconvex global objective functions, thus revealing its potential to be applied to a wider class of applications such as those using deep learning. In addition to desirable features that other state-of-the-art federated learning and swarm learning algorithms possess, DeceFL has additional desirable features beyond classical centralized federated learning and swarm learning, namely: 1) full decentralization: at any iteration, there is no central client that can receive all other clients' information, therefore avoiding data leakage; 2) principled: it has been proved that zero performance gap can be achieved when the loss function is strongly convex; 3) flexible communication topology design: any connected network structure suffices to achieve the training task; 4) all clients in the network can have the trained model, incentivizing clients to participate.

\section*{The proposed decentralized federated learning framework}
\subsection{Problem formulation}\label{sec:pf}
We first formulate the decentralized federated learning problem theoretically: assuming that there are $K$ clients with local data in the form for standard machine learning tasks: $\mathcal{D}_k$ for $k\in\{1,2,\ldots,K\}$. The training of AI models can be formulated as the following global learning problem (let $\mathcal{D}\triangleq \cup_k\mathcal{D}_k$ and $\cap_k\mathcal{D}_k=\emptyset$): $$\mathcal{M}_{\text{c}}\triangleq\arg\min_{\mathcal{M}} F(\mathcal{D}; \mathcal{M}).$$
Such an optimization problem cannot be directly solved without centralized information $\mathcal{D}$. However, clients would like to collaboratively train a model $\mathcal{M}_{\text{d}}$ using  the same objective function $F$, in which the $k$-th client does not send its data $\mathcal{D}_k$ to others. We define the performance gap as a nonnegative metric, which quantifies the degenerative performance between a centralized model and a decentralized one:
$$\Delta\triangleq F(\mathcal{D}; \mathcal{M}_{c}) - F(\mathcal{D}; \mathcal{M}_d).$$
The goal is to make $\Delta$ as small as possible, in the ideal case $\Delta=0$.
%
%

\subsection{The proposed DeceFL algorithm}

\begin{figure*}[!htb]
	\centering
	\includegraphics[width=\linewidth]{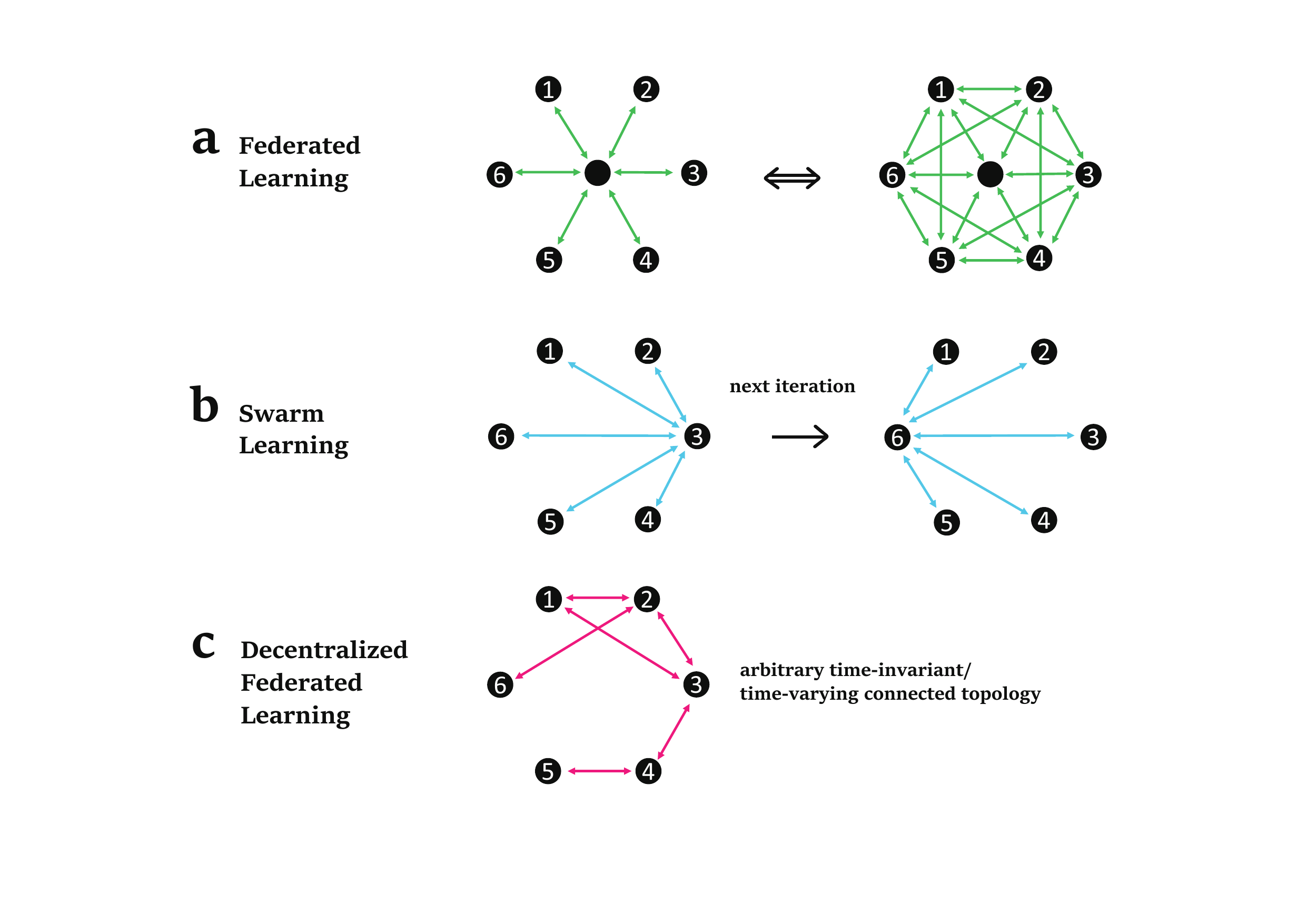}
\vspace{-3cm}\caption{{\bf Illustration of key concepts in different state-of-the-art federated learning frameworks.} {\bf a,} Classical Federated Learning: a central client is needed to receive and transmit all essential information to other clients. It is equivalent to an all-to-all network without such a central center, i.e., every client in the network can receive information from all other clients. {\bf b,} Swarm Learning: there is no such a universal central client, but a potentially different central client is selected in every iteration. Mathematically, it is equivalent to FedAvg with varying central clients. {\bf c,} The proposed Decentralized Federated Learning: there is no need for a central client in any iteration. Any connected time-invariant/time-varying topology would work, therefore unifying the classical federated learning and swarm learning.}
	\label{fig:main}
\end{figure*}

To solve the optimization problem in a decentralized way, we model the
communication network between clients as an undirected connected
\footnote{In this work, we consider that the information communication
  between clients is mutual for notational simplicity; therefore, the
  adjacency matrix $W=[W_{kj}] \in \mathbb{R}^{K\times K}$ is symmetric.
  Further we assume that the underlying topology is connected, i.e., for
  any two clients $k$ and $j$, there is at least one path from $k$ to
  $j$.} graph $\mathcal{G}=(\mathcal{N}, \mathcal{E}, W)$,
where $\mathcal{N} := \{1, 2, \ldots, K \}$ represents the set of clients,
and $\mathcal{E} \subseteq \mathcal{N} \times \mathcal{N}$
represents the set of communication channels, each connecting two distinct clients. For each edge $(i,j)\in \mathcal{E}$, the corresponding element in the adjacency matrix $W$, i.e., $W_{ij}$ indicates whether there is a communication channel between the $i$-th client and the $j$-th client. Specifically, when $W_{ij} > 0$, there is information communication between clients $i$ and $j$, while $W_{ij} = 0$ means none. For client $i$, when $W_{ij} > 0$, then client $j$ is called a neighbor of client $i$. The set of all such clients $j$ is represented as $\mathcal{N}_{i}$, i.e., $\mathcal{N}_{i} = \{ j | W_{ij} > 0, \forall j \in \mathcal{N}\}$. Define the local loss function $F_{k} (w) \triangleq F(\mathcal{D}_{k}; \mathcal{M})$ as the user-specified loss function on the dataset $\mathcal{D}_k$ with model parameters $w$ in $\mathcal{M}$, then $F(\mathcal{D}; \mathcal{M})$ can be rewritten as $F({w}) \triangleq \frac{1}{K}\sum_{k=1}^{K} F_{k} (w).$ Let the client $k$ hold a local copy of the global variable $w$, which is denoted by $w_{k} \in \mathbb{R}^{n}$, and $\mathbf{w} = [w_{1}; \ldots; w_{K}] \in \mathbb{R}^{K n}$.  Specifically, the update rule of DeceFL is, for each client $k=1, \ldots, K$,
\begin{align}\label{eq:DGD}
 {w}_{k}(t+1)
 = \underbrace{\textstyle\sum_{j=1}^{K} W_{kj} {w}_{j}(t)}_{\text{average of neighbors' estimates}}
  - \qquad \underbrace{\eta_{t} \nabla F_{k}({w}_{k}(t))}_{\text{gradient descent}},
 \end{align}
where $\eta_{t}>0$ is the learning rate, and the initial condition
$w_{k}(0) \in \mathbb{R}^{n}$ can be arbitrarily chosen.  Every client
is sharing with its neighbors (which is a subset of all other clients)
their model parameters rather than their data. Specifically, every client is running its local training algorithm, e.g., gradient descent, and it only communicates its own estimate of the global parameter with its neighbors. Once a client receives other estimates from neighboring clients, it averages out other estimates, adds to its local gradient and generates its estimate in the next iteration. The above process will be repeated until convergence.  As shown in Figure~\ref{fig:main}, in DeceFL, each client completes the update by receiving and transmitting directly with neighbor clients and local gradient calculation, without needing the aggregation and transmission of a third-party central client at any iteration. Thus, it is fully decentralized.

We stack the ${w}_{k}(t)$ and $\nabla F_{k}({w}_{k}(t))$ in \eqref{eq:DGD} into vectors, i.e., define $\mathbf{w}(t)
= [w_{1}(t)^T, \ldots, w_{K}(t)^T]^T \in \mathbb{R}^{K n}$ and $\nabla F (\mathbf{w}(t)) = [
\nabla F_{1}({w}_{1}(t))^T, \ldots,
\nabla F_{K}({w}_{K}(t))^T]^T \in \mathbb{R}^{K n}$. Then, we can compactly rewrite \eqref{eq:DGD} as
 \begin{align}\label{eq:DGD_2}
 \mathbf{w}(t+1) = (W\otimes \mathbf{I}_{n}) \mathbf{w}(t) - \eta_{t} \nabla F (\mathbf{w}(t) ),
\end{align}
where $W=[W_{ij}] \in \mathbb{R}^{K \times K}$ and $\mathbf{I}_{n} \in \mathbb{R}^{n\times n}$ is the identity matrix. Next, we analyze the convergence of DeceFL under the following assumptions about the global cost function, which is consistent with those made in the convergence analysis of FedAvg \cite{koloskova2019decentralized}.
\begin{assumption}
\label{assum:L_lipschitz}
For each $k = 1, \ldots, K$, assume that $F_{k}$ is $L_{k}$-smooth and $\mu_k$-strongly convex, where $L_{k}$, $\mu_k >0$.
That is, $F_{k}$ is differentiable and the gradient is $L_{k}$-Lipschitz continuous, i.e., for any $x, y\in \mathbb{R}^{n}$,
  \begin{align}\label{eq:Lipschitz1}
  \| \nabla F_{k}( x)-\nabla F_{k}(y) \|
  \leq
   L_{k} \|x-y\|,
  \end{align}
  and
   \begin{align}\label{eq:Lipschitz2}
  F_{k}(x) \geq F_{k}(y)
  + \langle \nabla F_{k}(y), x-y \rangle
+ \frac{\mu_{k}}{2} \|x-y\|^2.
  \end{align}
\end{assumption}
When Assumption~\ref{assum:L_lipschitz} holds, the global objective function ${F}(\cdot)$
is $L$-smooth and $\mu$-strongly convex, where $L = \max\{ L_{1}, \ldots, L_{K}\}$ and $\mu = \min\{\mu_{1}, \ldots, \mu_{K}\}$. Clearly, $\mu \leq L$.

Assumption \ref{assum:L_lipschitz} is standard and satisfied by typical loss functions in machine learning, including $l_2$-regularized linear regression and  $l_2$-regularized logistic regression. In order to analyze the convergence of the algorithm, we define the average sequence $\bar{w}(t) = \frac{1}{K}(\mathbf{1}^{T}_K\otimes \mathbf{I}_{n})\mathbf{w}(t) = \frac{1}{K} \sum_{k=1}^{K} w_{k}(t)
$, where $\mathbf{1}_K \in \mathbb{R}^{K}$ is a vector in which all elements are 1. Denote $\lambda$ as the spectral norm of $W - \frac{1}{K} \mathbf{1} \mathbf{1}^T$, where $\lambda\in (0, 1)$ from algebraic graph theory (Supplementary Information).

 \begin{theorem}
\label{Th:DGD_FL_adaptive}
Consider algorithm \eqref{eq:DGD}, where the learning rate chosen by $\eta_t = \frac{\delta}{t+\Gamma}$, in which $\delta > \frac{1}{\mu}$ and $\Gamma > {\frac{\lambda}{1 - \lambda} }$ satisfying $\frac{\delta }{\Gamma} \leq  \frac{1}{L}$. Denote the gap between the average local optimal value and the local function value at the initial point $\mathbf{w}(0)$, as $\varepsilon_{0} \triangleq \sum\nolimits_{k=1}^{K} (F_k ({w}_k(0)) - F_k ({w}^{\ast}_k )) \geq 0$, where ${w}^{\ast}_k =\arg\min_{w_k}F_k(w_k) $.  Then the following inequality can be obtained under Assumption \ref{assum:L_lipschitz}:
  \begin{align}
 \label{eq:th_client_avg_error}
\|\mathbf{w}(t) - \mathbf{1} \otimes \bar{w}(t) \| \leq \frac{\zeta}{t+\Gamma},
 \end{align}
 and
   \begin{align}
 \label{eq:th_avg_error}
\|\bar{w}(t) - w^{\ast} \|
\leq
\frac{\tilde{\zeta}}{t+\Gamma},
 \end{align}
 where
 $\zeta \triangleq \max\bigg\{ \Gamma \|\mathbf{w}(0) - \mathbf{1} \bar{w}(0) \| , \frac{\delta \sqrt{2L\varepsilon_{0}  }}{{\frac{\Gamma}{\Gamma +1 } - \lambda}}\bigg\}$
and
${\tilde{\zeta}} \triangleq\max\big\{
\Gamma \|\bar{w}(0) - w^{\ast} \|, \
\frac{1}{\mu \delta -1}\frac{L \delta \zeta}{\sqrt{K}}
 \big\}$.
\end{theorem}
\bproof
See Supplementary Information.
\eproof

In \cite{li2019convergence}, the authors studied the convergence of FedAvg and established the convergence rate $O(1/T)$ (where T is the number of iterations in gradient descent) for strongly convex and smooth problems. Theorem~\ref{Th:DGD_FL_adaptive} guarantees the convergence of the proposed DeceFL algorithm under time-invariant connected communication topologies with the same convergence rate. In the Supplementary Information, it has been further demonstrated that the proposed DeceFL converges even for time-varying topologies, in which case the underlying topology does not need to be connected for all iterations.

\subsection{An illustrative example}
We use a simple yet important example to illustrate the problem and demonstrate the advantages of the proposed decentralized framework as compared with centralized learning, federated learning and swarm learning (SL). Consider the case where $K$ clients would like to compute the average of every client's private value $\mathcal{D}_k=w_k(0)$, i.e., $w^*\triangleq\frac{\sum_{k=1}^K w_k(0)}{K}$. This task is simple if there exists a central client. However, it becomes challenging when a central client is not available. This is the well-known decentralized consensus problem \cite{olfati2004consensus, ren2005consensus,jadbabaie2003coordination}, which has many engineering applications including synchronization, PageRank, state estimation, load balancing, and more.

We can convert the consensus problem to the following optimization problem
\begin{align}
   \label{eq:obj_consensus}
     \min_{w \in \mathbb{R}} F(\mathcal{D}; {w})
     \triangleq
    \frac{1}{2} \sum^{K}_{k=1} (w-w_k(0))^2,
   \end{align}
with its optimal value coincided with $w^*$. Rather than computing the mean, we convert it to solve the optimization problem \eqref{eq:obj_consensus} using algorithms FedAvg, SL and the proposed DeceFL respectively.

\subsubsection{FedAvg algorithm}
In the classical federated learning algorithm, i.e., FedAvg, there is a central server to collect the local parameter information of each client for average aggregation, and to assign it to each client, which is equivalent to a complete graph of $K$ clients without a central server after simple derivation (Figure~\ref{fig:main}A):
  \begin{equation}
   \label{eq:FedAvg}
   w_{k}(t+1) = \frac{1}{K} \sum_{k =1}^{K} w_{k}(t) - \eta_t(w_{k}(t) - w_k(0)),
   \end{equation}
where $t$ represents the $t$-th iteration, $w_k(t)$ represents the estimate of global optimum for the $k$-th client at iteration $t$, and $\eta_t$ is learning rate in the gradient descent algorithm. In essence, every client iteratively updates its estimate based on all others' estimates together with its current gradient. Using derivation in the Supplementary Information  based on dynamical system and algebraic graph theory, it can be shown that the system reaches the steady-state, i.e., $\lim_{t\rightarrow\infty}\|w_k(t)-w^*\|=0$, if $\eta_t =\frac{\gamma}{t+\Gamma}$ and $ \frac{\gamma}{\Gamma} < 1$ is satisfied for any ${\gamma}, {\Gamma}>0$.

\subsubsection{SL algorithm}
The SL algorithm considers the situation where there is no central server: in each iteration, a random leader is dynamically selected from the members to aggregate the model parameters from all clients (including itself) and assign them to each client shown in Figure~\ref{fig:main}B. Mathematically, this is exactly the same as centralized federated learning as shown in the Supplementary Information.

\subsubsection{DeceFL algorithm}
Each client communicates parameters through the topology of the undirected connected graph shown in Figure~\ref{fig:main}C. According to the weighted aggregation of information obtained from neighbor clients, the local update is completed according to the following iteration,
  \begin{align}
   \label{eq:DeceFL}
   w_{k}(t+1)
 = \sum_{j =1}^{K}{W_{kj}(t)}w_{k}(t) - \eta_t (w_{k}(t) - w_k(0)),
   \end{align}
where $W(t) \in \mathbb{R}^{K\times K}$ is the weighted matrix of the undirected connected graph at iteration $t$. Using derivation in the Supplementary Information  based on dynamical system and algebraic graph theory, it can be shown that the system reaches the steady-state, i.e.,  $\lim_{t\rightarrow\infty}\|w_k(t)-w^*\|=0$, if $\eta_t =\frac{\gamma}{t+\Gamma}$ and $ \frac{\gamma}{\Gamma} < 1-\sigma'$\footnote{Let us sort the eigenvalues of $W$ in a non-increasing order as
$1 = \lambda_{1}(W) >  \lambda_{2}(W) \geq \cdots \geq  \lambda_{n}(W) > -1$, denoted $\sigma'$ as the second largest eigenvalue of the weighting matrix $W$, i.e., $\sigma' = \max\{| \lambda_{2}(W)|, | \lambda_{n}(W)|\} \in(0, 1)$.} are satisfied for any ${\gamma}, {\Gamma}>0$.

\subsubsection{Summary}
It can be shown that all methods can reach consensus with zero performance gap, i.e., $\Delta=0$. The convergence speeds of all methods are $O(1/T)$. As shown in the Supplementary Information Figure~1, these methods converge to the consensus value exponentially. This is consistent with the theoretical results. The information used in three different algorithms is however distinct: FedAvg and SL need global information at every iteration, while DeceFL only needs local information from clients' neightbors. In addition, the first two algorithms can be viewed as special cases of the proposed DeceFL algorithm by setting $W$ to the corresponding adjacency matrix correspondingly. Specifically, when $W(t)=\frac{1}{k}\textbf{1}_K\textbf{1}_K^T$ for all $t$, the proposed DeceFL algorithm becomes to FedAvg and SL. Therefore, the proof of convergence in this paper also warrants that of centralized federated learning and SL.

\section*{Experiments}

Experiments were carried out on real-world biomedical and industrial
applications, which demonstrate the effectiveness and the wide
applicability of DeceFL, as a fully decentralized framework. We
benchmarked the performance of DeceFL, in comparison with FedAvg and SL
that demand much more communication costs and strongly reply on a
restricted communication topology. Furthermore, the superiority of DeceFL
on robustness in the presence of communication topology interference (random node
or edge malfunction) was shown in two experiments with time-varying communication topologies. The overall performance of DeceFL was corroborated by these practical applications.

\setcounter{subsection}{0}

\subsection{Application to medicine: collaborative prediction of leukaemias}

First, we used the dataset of peripheral blood mononuclear cell (PBMC)
transcriptomes from \cite{2021Nature_SL}, named ``dataset A2'', as a
benchmark example to compare three federated learning frameworks:
DeceFL, FedAvg and SL. Samples were split into non-overlapping training
datasets associated with each node, and a global test dataset that was
reserved for testing the models built on these frameworks.
The experiment setup is consistent with that of SL: the dataset is
divided into a training set and a test set at the ratio of 8:2, and the
dataset owned by each node was obtained from the training set. The
logistic regression model with $l_2$ regularization and the 8-layer
fully connected deep neural network (as in \cite{2021Nature_SL}) were
selected for our experiments detailed in the Materials and Methods
Section.

 \begin{figure*}[!htb]
	\centering
	\includegraphics[width=\linewidth]{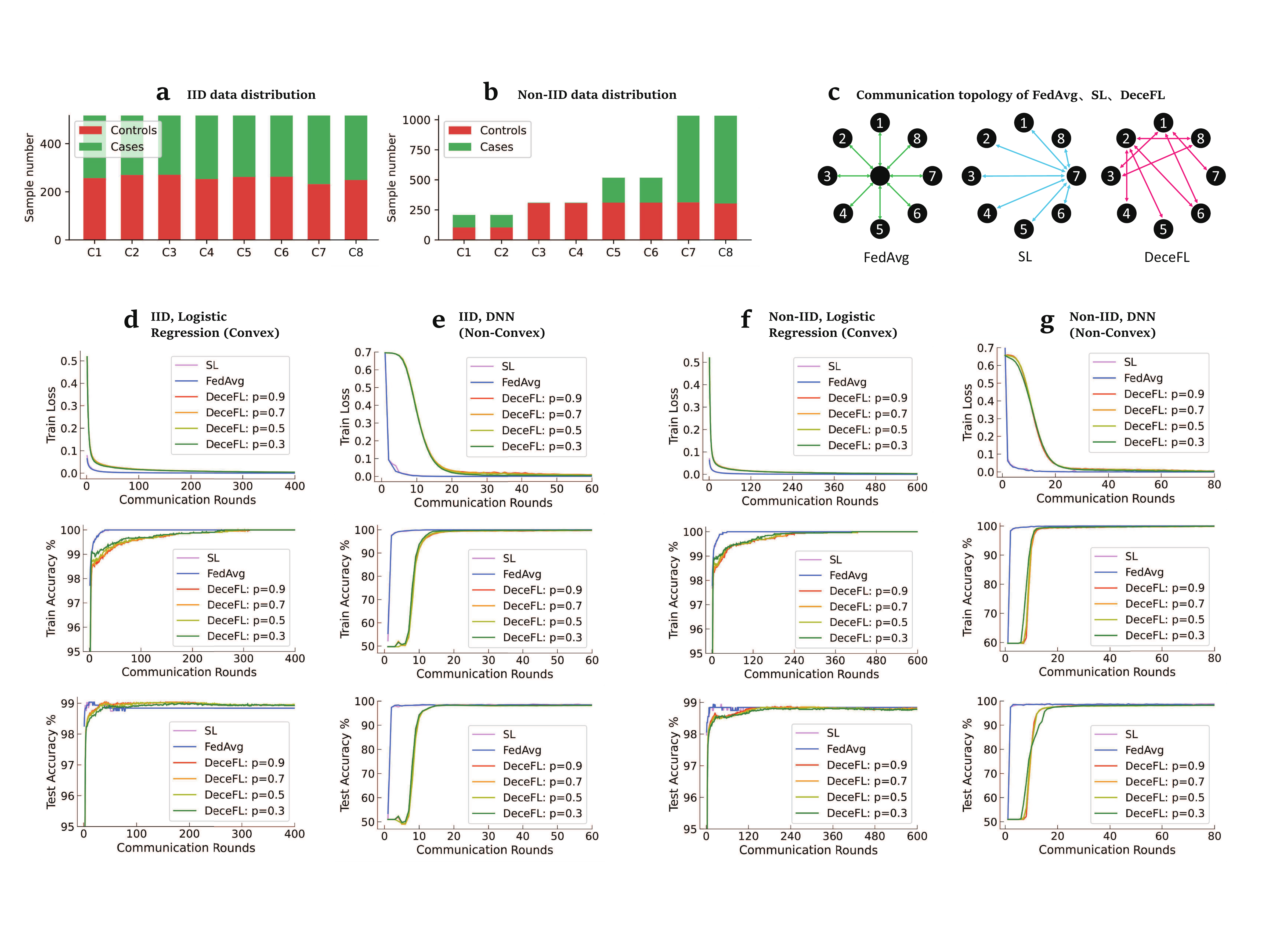}
    \vspace{-2cm}\caption{{\bf DeceFL to predict leukaemias from A2
        benchmark dataset \cite{2021Nature_SL}.}
      {\bf a,} Data were divided into IID samples for all clients.
      {\bf b,} Data were divided into Non-IID unbalanced samples.
      {\bf c,} Different topologies for FedAvg, SL and DeceFL
      respectively. The topology for SL must hold in every iteration
      when any other node is selected as a central client.
      {\bf d, e, f, g,} Performance of three algorithms on IID/Non-IID
      setups over logistic regression/neural networks.}
	\label{fig:A2}
\end{figure*}

First, we benchmarked DeceFL against FedAvg and SL in the IID setup of
dataset A2 (Fig.~\ref{fig:A2}a), that is, the sample size of each node
is the training set sample size divided by the number of nodes, which
ensures that each node has the same number of samples and the ratio of
the positive to the negative samples is approximately $1:1$. DeceFL
applied multiple connected graphs with various connectivity probability
values ($p = 0.3, 0.5, 0.7, 0.9$). This benchmark shows that DeceFL can
reach the same performance as FedAvg and SL which use a (temporary)
central client to gather all information from every node. FedAvg and SL
only perform better during the transient period that DeceFL takes for a
certain number of iterations to converge due to its decentralized
nature. Second, the similar comparative study was repeated with the
Non-IID setup of dataset A2 (Fig.~\ref{fig:A2}b). The Non-IID setup
explicitly designs, for the local data associated with each node, the
sample size and the ratio between positive and negative samples
(Supplementary Information). It allows us to benchmark performances on
balanced/unbalanced, sufficient/deficient local training data. We
obtained very similar results as in the IID setup, where DeceFL presents
an equal performance to FedAvg and SL, after DeceFL reaches consensus in
decentralized computation. It also shows the superiority of DeceFL over
SL, which however demands huge amounts of communication costs for the
selected central client at every round and relies heavily on the strong
assumption of a stable fully-connected communication structure. Any bit
of malfunction of clients or communication paths could melt down the
whole SL process, since at each round a client is delegated to collect
information from all other clients.

\begin{figure*}[!htb]
	\centering
	\includegraphics[width=\linewidth]{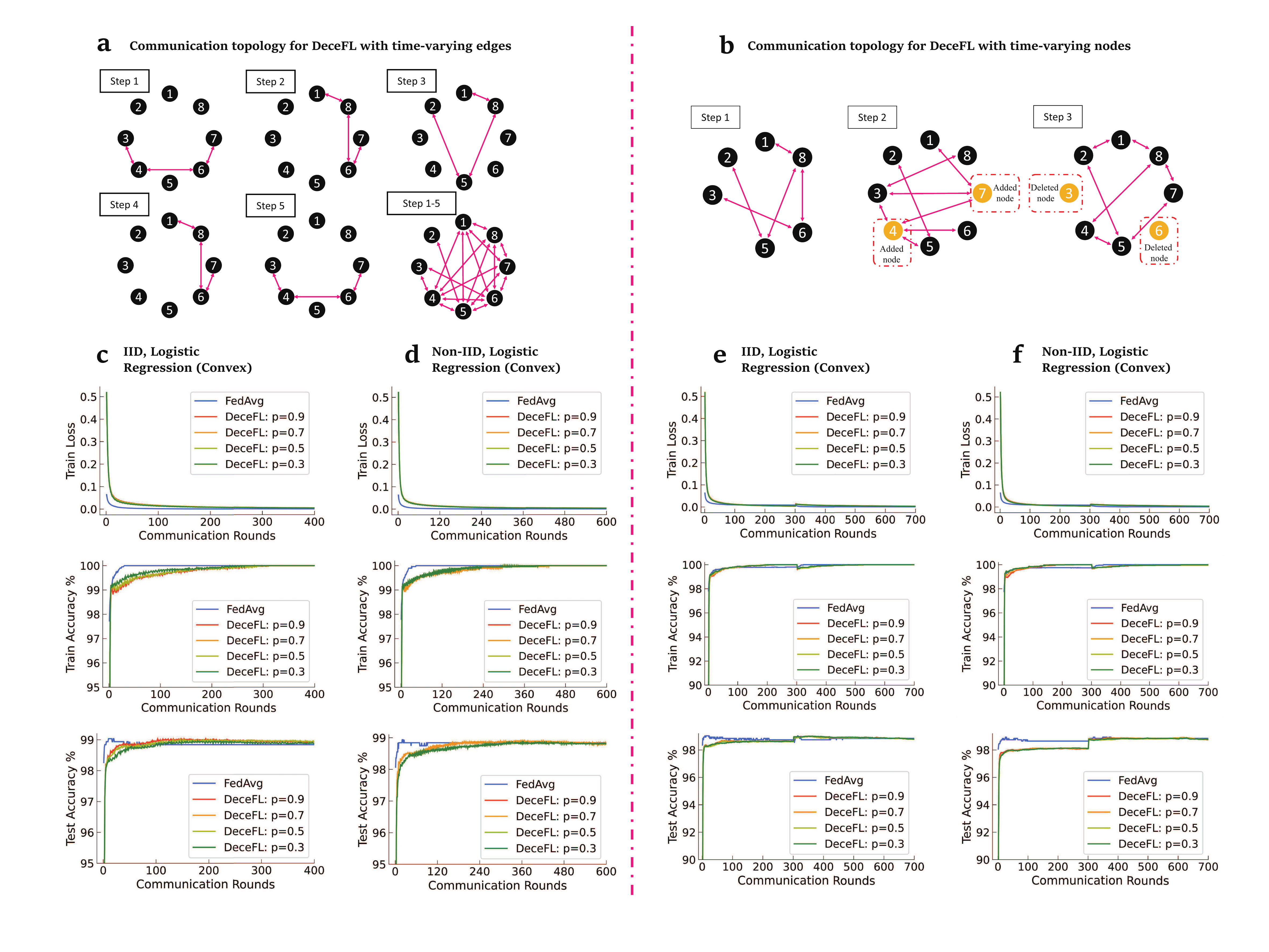}
    \vspace{-2cm}\caption{{\bf DeceFL to predict leukaemias from A2 benchmark dataset \cite{2021Nature_SL}.}
      \textbf{a,} Time-varying communication topology that consists of a sequence of graphs each of which is not connected while the lump-sum graph over a fixed period is connected.
      \textbf{b,} Time-varying communication topology that adds or removes nodes over time.
      \textbf{c,d,} Performance of DeceFL with edge-varying graphs on the IID and Non-IID setups of dataset A2 using logistic regression, with reference performance of FedAvg that uses full information.
      \textbf{e,f,} Performance of DeceFL with node-varying graphs on the IID and Non-IID setups of dataset A2 using logistic regression, with reference performance of FedAvg that uses full information.}
      \label{fig:A22}
\end{figure*}

To show DeceFL functions well when an intervention to decentralized
infrastructure happens, we conducted two experiments with time-varying
graphs that take into account malfunction of clients and communication
paths. First, communication structure/graph was altered in runtime
(Fig.~\ref{fig:A22}a), that is, the adjacency matrix that describes how
nodes communicate with each other varied over time. Although being named
as a decentralized framework, SL requires a fully-connected
communication graph; whereas DeceFL only demands connected graphs as
shown in the IID and Non-IID experiments. This time-varying experiment
further shows that the conditions of DeceFL can even be weaken and
generalized to such an extent that the communication graph at each time
is not necessary to be connected as long as within a fixed period the
information can be transmitted between any pair of nodes. Surprisingly,
both experimental results Fig.~\ref{fig:A22}c (IID) and
Fig.~\ref{fig:A22}d (Non-IID) show that DeceFL in such a scenario keeps
similar performance as FedAvg. In other words, DeceFL can be so robust
that random malfunction of a small portion of edges may merely
deteriorate DeceFL running processes. The second experiment considers
the removal and supplement of nodes, as shown in Fig.~\ref{fig:A22}b:
during the first 300 rounds, we used an Erdos-Renyi graph of 6 nodes;
then for the 301-600 rounds, the graph was of 8 nodes by adding 2 extra
nodes; and in the rest rounds, the graph was randomly removed by 2
nodes. Under such node interventions, experimental results
Fig.~\ref{fig:A22}e and Fig.~\ref{fig:A22}f show robust performance of
DeceFL which is similar to FedAvg (that does not consider node
interventions). Two experiments manifest the robustness of DeceFL on
interventions of computation infrastructure in a decentralized
framework.

\subsection{Application to smart manufacturing: collaborative detection of bearing faults}
\label{sec:cd-dataset}
Modern manufacturing is heavily influenced by AI technologies with
extraordinary increase of computational power and data size. To raise
productivity and reduce operational costs, a critical challenge is fault
diagnosis in machining operations \cite{isermann2005fault}. AI-based
algorithms have the potentials to detect fault locations and even to
predict faults in advance, which allow replacing regular maintenance
with real-time data-driven predictive maintenance and further reduce
unnecessary maintenance costs and guarantee reliability. A general fault
detection framework has been proposed in \cite{yuan2020general}, which,
however, needs full-cycle measurements of large amounts of machines that
are most likely unavailable from a single factory. Data generated by
multiple factories could be sufficient to perform preventive
maintenance, while sensitive data (security or business related) are
less likely to be shared in practice. The fully decentralized framework
DeceFL provides a way for multiple factories to develop a global model,
which generates mutual benefit from private local data without having
to resort to data sharing in public.

This experiment practices such a decentralized fault diagnosis
application in manufacturing, using Case Western Reserve University's
(CWRU) bearing data, which comprises ball bearing test data for normal
and faulty bearings, specified in the Methods Section. Specifically, we
used three types of bearings data: 7 inch, 14 inch and 21 inch; and
chose the drive end defects, which includes outer race defect, inner
race defect, and ball defect. We chose the outer race defect appearing
at the 6 o'clock (centered) position. Thus, there are in total ten
distinct conditions: 9 faulty classes (3 bearing types times 3 defect
types) and the normal condition. All data in use was collected at 12,000
samples/second for drive end bearing experiments. It was generated by using 4
types of motor speed: 1797 rpm, 1772 rpm, 1750 rpm and 1730 rpm.
The data from 1730 rpm is reserved for test.

\begin{figure*}[!htb]
	\centering
	\includegraphics[width=\linewidth]{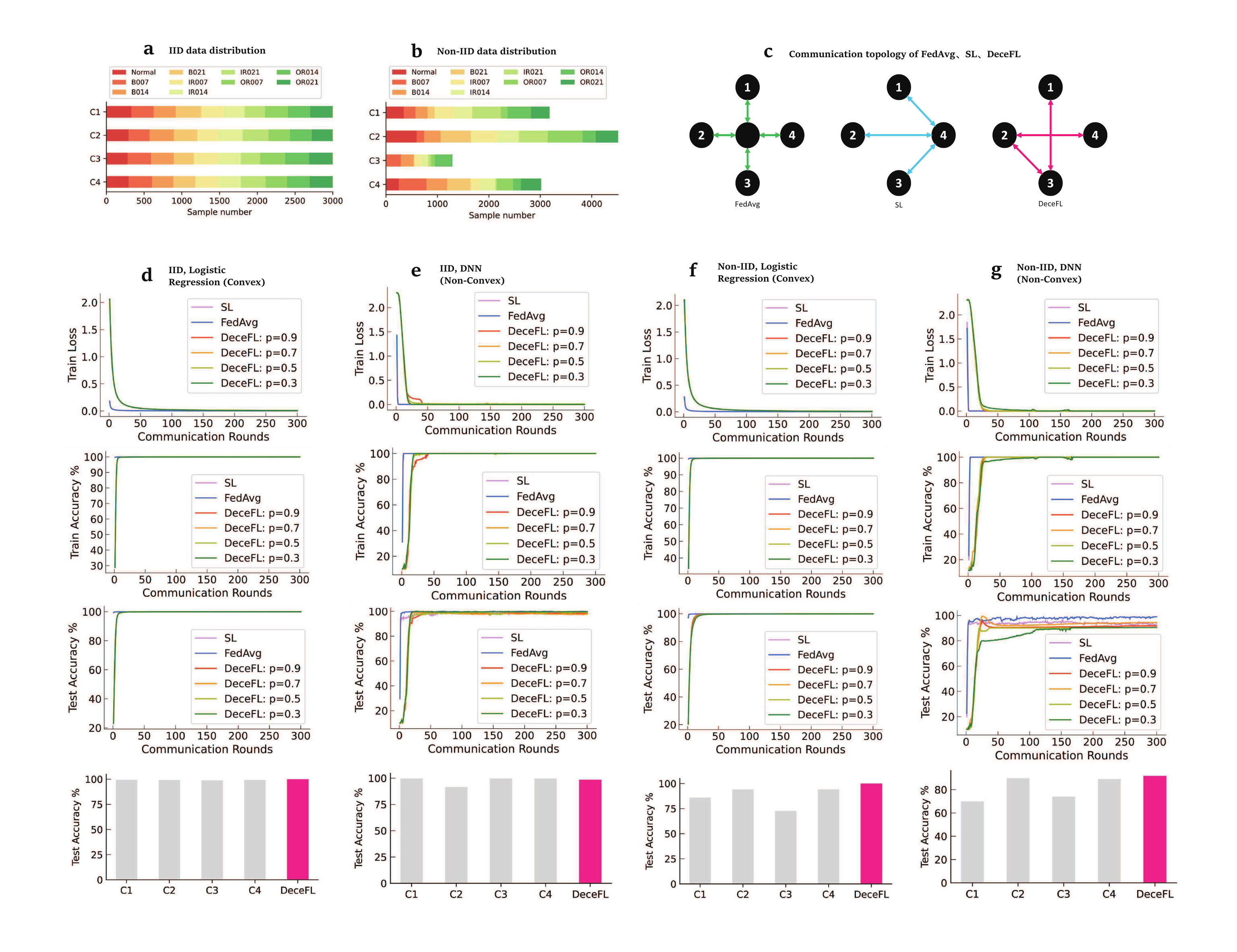}
   \vspace{-2cm} \caption{{\bf DeceFL to detect bearing faults from CWRU benchmark dataset.}
     \textbf{a,} Data were divided into IID samples for all 4 clients.
      \textbf{b,} Data were divided into Non-IID unbalanced samples.
      Each client locally specified its data size and sample distribution.
      \textbf{c,} Illustration of communication topology for FedAvg, SL and DeceFL.
      \textbf{d,e,} Performance of DeceFL on IID data using logistic regression and DNN, respectively, with reference performance of FedAvg and SL.
      \textbf{f,g,} Performance of DeceFL on Non-IID data using logistic regression and DNN, respectively, with reference performance of FedAvg and SL.
      \textbf{d,e,f,g,} The boxplots at bottom illustrate the
      performance comparison between DeceFL and each client trained
      independently (that is, each client trained its own model only using its
      associated local data without communicating with any other clients).
      }
	\label{fig:CWRU}
\end{figure*}

Assume that there are 4 factories, as clients illustrated in
Fig.~\ref{fig:CWRU}c, which collect their private full-cycle bearing
data. The training data associated with each client were prepared in the
IID (Fig.~\ref{fig:CWRU}a) and the Non-IID setup (Fig.~\ref{fig:CWRU}b).
A 10-way classification problem is considered, 9 fault cases (B007,
IR007, OR007, B014, IR014, OR014, B021, IR021, OR021) and 1 normal case.
Learning used two methods, regularized logistic regression as a strongly
convex method, and deep neural network (DNN) as a nonconvex method.
In the usage of logistic regression, as guaranteed in theory, DeceFL in
Fig.~\ref{fig:CWRU}d,f confirms the same performance as FedAvg after its
transient periods. For the case of DNN, as a non-convex method, although
there is no theoretical guarantee, DeceFL in Fig.~\ref{fig:CWRU}e,g
shows competitive performance to FedAvg. The slight performance gap in
test between DeceFL and FedAvg in Fig.~\ref{fig:CWRU}f may be mostly
caused by the chosen type of DNN, multilayer perceptrons as used in
\cite{2021Nature_SL}, which has many well-known defects; and more
reasons are discussed and explored by more experiments in the
Supplementary Information. Comprehensive experiments were conducted and
can be found in the Supplementary Information, with more clients (that
is, more factories), more learning methods, and another 4-way
classification problem. Overall DeceFL manifests competitive performance
on multi-class classification for industrial fault diagnosis
applications, with implementations of (non-)convex methods in a fully
decentralized framework that breaks through the barrier of data privacy.

\section*{Discussions}
\label{section:Future}
In this paper, we propose a new decentralized federated learning algorithm. The decentralized architecture eliminates the bandwidth bottleneck of the central client. The convergence of the DeceFL algorithm is analyzed in detail, showing that DeceFL guarantees convergence and has the same convergence rate as the centralized federated learning algorithm. The convergence performance of the algorithm is verified by training neural networks over different datasets. Compared with other state-of-the-art privacy-preserving algorithms such as FedAvg and SL \cite{2021Nature_SL}, the proposed DeceFL algorithm is guaranteed to reach the global optimum with a similar rate as the centralized federated learning algorithm under certain conditions. In addition, there has developed a sizable literature as surveyed in \cite{2019ARC}, which can be adapted to cope with quantization errors and noises that could happen over communication networks.

There is no doubt that decentralized federated learning framework will become increasingly popular in the nearest future for almost all AI applications given the privacy regulations. Yet our algorithm has a number of limitations that need to be taken into consideration for future development:

First of all, application of privacy algorithms (for example, blockchain or homomorphic encryption \cite{demillo1978foundations}) has not been considered in this study. However, similar to the centralized federated learning and swarm learning framework, it should be straightforward to apply such techniques for data privacy protection in the proposed DeceFL framework to make communication secure.

Secondly, similar to the setup of federated learning and swarm learning, all clients in the network need to know the form of the global objective function. The proposed formulation is different from those used in multi-party computation \cite{yao1982protocols}, where model and data can be separated \cite{yu2021privacy}. Future work lies in the integration of such techniques to the proposed DeceFL algorithm to make the global objective function unknown to clients.

Finally, all clients are assumed to be collaborative, it would be interesting to further investigate whether the proposed decentralized federated learning framework is vulnerable to semi-honest or malicious clients that are not collaborative \cite{lyu2020threats}, which could exist in the real-world applications.

\section*{Materials and Methods} \small

\hspace*{-15pt}\textbf{Data pre-processing} \\
Essential data preprocessing was performed for CWRU dataset, including
class balancing and normalization, feature extraction by Fourier
transform. The dataset has 10 classes in total, which vary in sample
size. Hence samples in certain classes were deleted to balance sample
sizes over all classes. The original data is time-series data, which was
firstly divided by every 300 points and resulted in a family of time
series. Each time series chose DE and FE features respectively, and
produced 600 points. For every time series of each feature, we performed
Fast Fourier Transform (FFT), which yielded 150 points. Thus each time
series of both DE and FE has in total 300 points. The motivation to use
FFT is to handling the mismatch of time stamps of sequential data. After
FFT the training and test data were then normalized by removing the mean and
scaling to unit variance (the test data is normalized by the normalizer
of the training).

\hspace*{-15pt}\textbf{Performance metrics} \\
The common performance metric ``accuracy'' is used for assessment of
classification,
\begin{equation}
\text{accuracy} = \frac{\text{TP}+\text{TN}}{\text{TP}+\text{TN}+\text{FP}+\text{FN}}
\end{equation}
where TP, TN, FP and FN denote the number of true positive, true negative, false positive and false negative samples, respectively.

\hspace*{-15pt}\textbf{Implementation of learning methods} \\
To ensure in benchmark DeceFL can work well for either strongly convex
learning methods or non-convex methods, we adopted two algorithms:
logistic regression with $l_2$ regularization, and deep neural
network (DNN), as used in \cite{2021Nature_SL}.
For logistic regression, every node runs 10 epochs in each
round, with batch-size $64$. It uses the SGD optimizer, with weight
decay coefficient $10^{-4}$ for the realization of $l_2$ regularization.
The initial learning rate (for deep learning framework) is $0.01$, which
is later decayed by multiplying $0.2$ every $5$ epochs.
For DNN, every node runs 30 epochs in each round, with
batch-size $64$. It uses the SGD optimizer, with weight decay
coefficient $10^{-4}$. The initial learning rate (for deep learning
framework) is $0.1$, which is decayed by multiplying $0.2$ every $20$
epochs. This DNN has $8$ hidden layers, whose dimensions are
$256, 512, 512, 256, 256, 128, 128, 64$, respectively. The
\emph{dropout} rate is set to $0.3$.
Both methods use \emph{sigmoid} as the activation function in the output
layer for binary classification (dataset A2) and \emph{softmax} for
multiclass classification (CWRU dataset). At aggregation, the gradient
update coefficient for DeceFL is $0.1$ (FedAvg does not use this variable).
The total number of running rounds is selected by visualization effects
of convergence for all methods in comparison.


\hspace*{-15pt}\textbf{Data availability}\\
The peripheral blood mononuclear cell (PBMC)-derived transcriptome
dataset, named as ``dataset A2'' in \cite{2021Nature_SL}, was used,
which was originally generated with Affymetrix HG-U133 2.0 microarrays
(8,348 individuals), by inspection of all publicly available datasets at
National Center for Biotechnology Information Gene Expression Omnibus.
To perform the IID experiments, the initial preparation of dataset randomly
dropped negative samples, resulting in 5,176 samples, such that the
whole dataset is balanced, i.e. the ratio of the positive to the
negative samples is $1:1$. CWRU Bearing dataset refers to the data of
ball bearing test for normal and faulty bearings from the Case Western
Reserve University, available on
\url{https://engineering.case.edu/bearingdatacenter}.

\hspace*{-15pt}\textbf{Code availability}\\
All source codes are openly available on GitHub \scriptsize (\url{https://github.com/HAIRLAB/DeceFL}).\small

\hspace*{-15pt}\textbf{Acknowledgement}\\
We thank Mr. Anthony Haynes, Mr. Cai Huang for editing.

\hspace*{-15pt}\textbf{Funding}\\
This work was supported by Jiangsu Industrial Technology Research Institute (JITRI) and the National Key R\&D Program of China (2018YFB1701202).

\hspace*{-15pt}\textbf{Author contributions}\\
Idea was conceived by Y.Y.. Theory was developed by J.L., R.C., L.X., X.Y., T.Y.,Y.Y.. Simulation codes were developed by D.J., C.S., M.W. and were reviewed by Z.Y.. Experiments were designed by Y.Y., Z.Y., and were performed by D.J., Z.Y., C.S., M.W., Y.Y., F.H., R.C.. Projects were supervised by Y.Y., S.S., H.D.. Funding was acquired by Y.Y., J.L., H.Z., H.D.. The original draft was written by Y.Y., Z.Y., J.L., R.C., and all authors provided critical review of the manuscript and approved the final draft.

\hspace*{-15pt}\textbf{Competing interests}\\
The authors declare no competing interests.

\vspace{5mm}
\scriptsize
\bibliography{YY}

\onecolumn \normalsize

\includepdf[pages={1-43}]{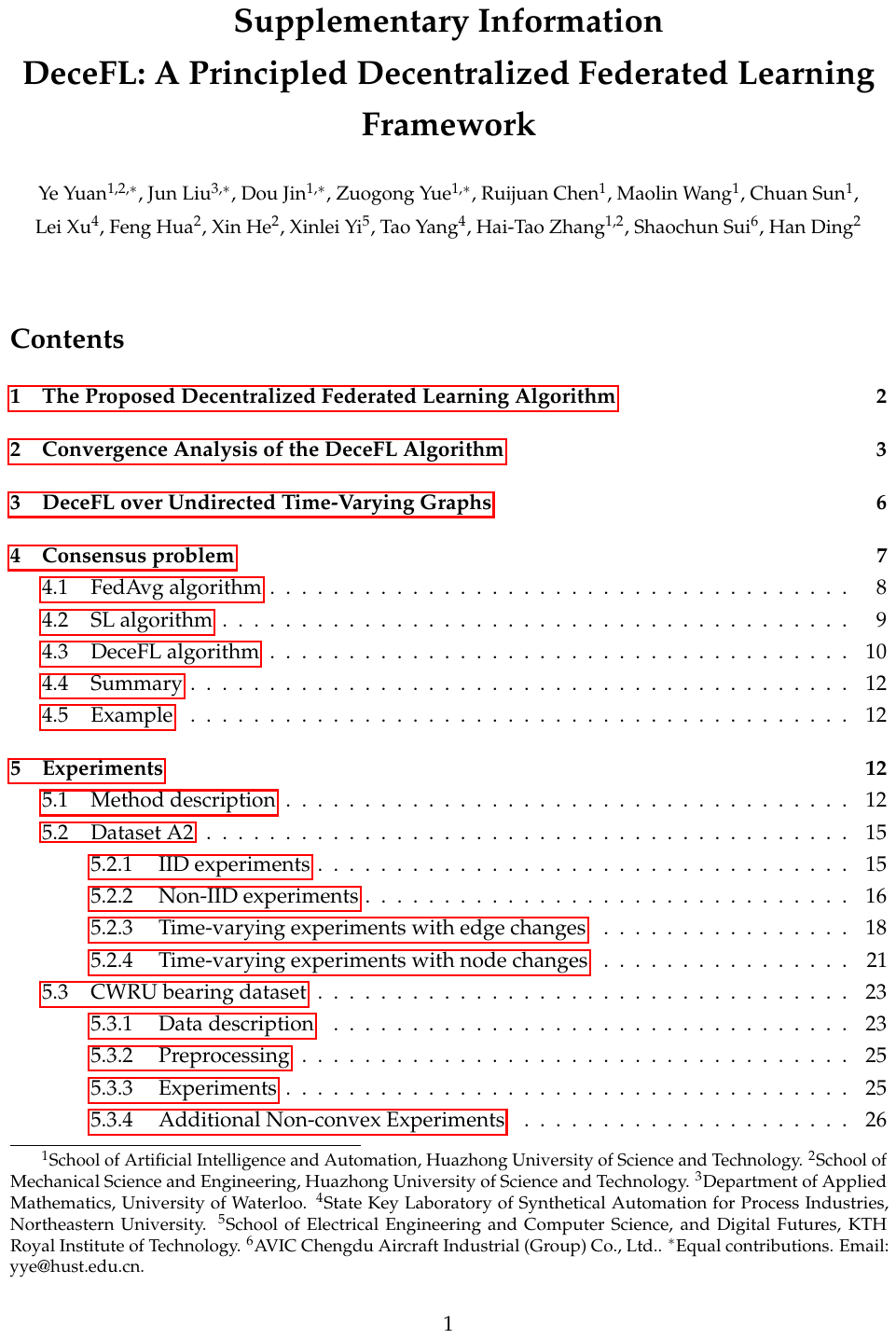}

\end{document}